\documentclass{article}

\usepackage{microtype}
\usepackage{graphicx}
\usepackage{subcaption}
\usepackage{booktabs} %

\usepackage{hyperref}

\usepackage[preprint]{icml2026}

\usepackage{amsmath}
\usepackage{amssymb}
\usepackage{mathtools}
\usepackage{amsthm}
\usepackage{array}
\usepackage{booktabs}
\usepackage{graphicx}
\usepackage{pifont}
\usepackage{colortbl}      
\usepackage{algorithm}
\usepackage{algorithmic}
\usepackage{multirow}
\usepackage{xspace}
\usepackage[capitalize]{cleveref}

\makeatletter
\DeclareRobustCommand\onedot{\futurelet\@let@token\@onedot}
\def\@onedot{\ifx\@let@token.\else.\null\fi\xspace}
\def\eg{\emph{e.g}\onedot}
\def\ie{\emph{i.e}\onedot}

\def\vs{\emph{vs}\onedot}

\makeatother
\usepackage{pifont}
\newcommand{\cmark}{\ding{51}}  %
\newcommand{\xmark}{\ding{55}}  %

\theoremstyle{plain}

\theoremstyle{definition}

\theoremstyle{remark}

\usepackage[disable,textsize=tiny]{todonotes} 

\icmltitlerunning{PromptRL: Prompt Matters in RL for Flow-Based Image Generation}

\begin{document}

\twocolumn[
  \icmltitle{PromptRL: Prompt Matters in RL for Flow-Based Image Generation}

  \icmlsetsymbol{equal}{*}

  \begin{icmlauthorlist}
    \icmlauthor{Fu-Yun Wang}{CUHK,Reve,equal}
    \icmlauthor{Han Zhang}{MSL}
    \icmlauthor{Michaël Gharbi}{Reve}
    \icmlauthor{Hongsheng Li}{CUHK}
    \icmlauthor{Taesung Park}{Reve}
  \end{icmlauthorlist}
  
  \icmlaffiliation{Reve}{Reve, USA}
  \icmlaffiliation{CUHK}{The Chinese University of Hong Kong, Hong Kong}
  \icmlaffiliation{MSL}{Meta Superintelligence Labs, USA}

  \icmlcorrespondingauthor{Han Zhang}{hanzhang.ai@gmail.com}
  \icmlcorrespondingauthor{Hongsheng Li}{hsli@ee.cuhk.edu.hk}

  \icmlkeywords{Flow Matching, Reinforcement Learning}

  \vskip 0.3in
]

\printAffiliationsAndNotice{\textsuperscript{*}Work done during an internship at Reve.}

\begin{abstract}
    Flow matching models (FMs) have revolutionized text-to-image (T2I) generation, with reinforcement learning (RL) serving as a critical post-training strategy for alignment with  reward objectives. In this research, we show that current RL pipelines for FMs suffer from two underappreciated yet important limitations:  sample inefficiency due to insufficient generation diversity, and pronounced prompt overfitting, where models memorize specific training formulations and exhibit dramatic performance collapse when evaluated on semantically equivalent but stylistically varied prompts. We present \textbf{PromptRL} (\textbf{P}rompt \textbf{M}atters in \textbf{RL} for Flow-Based Image Generation), a framework that incorporates language models (LMs) as trainable prompt refinement agents directly within the flow-based RL optimization loop. This design yields two complementary benefits: rapid development of sophisticated prompt rewriting capabilities and, critically, a synergistic training regime that reshapes the optimization dynamics. PromptRL achieves state-of-the-art performance across multiple benchmarks, obtaining scores of 0.97 on GenEval, 0.98 on OCR accuracy, and 24.05 on PickScore. 
    Furthermore, we validate the effectiveness of our RL approach on large-scale image editing models, improving the EditReward of FLUX.1-Kontext from 1.19 to 1.43 with only 0.06 million rollouts, surpassing Gemini 2.5 Flash Image (also known as Nano Banana), which scores 1.37, and achieving comparable performance with ReasonNet (1.44), which relied on fine-grained data annotations along with a complex multi-stage training.
Our extensive experiments empirically demonstrate that PromptRL consistently achieves higher performance ceilings while requiring over 2$\times$ fewer rollouts compared to naive flow-only RL. Our code is available at \url{https://github.com/G-U-N/UniRL}.
\end{abstract}

\section{Introduction}
\label{sec:intro}

\begin{figure*}[t]
    \centering
    \includegraphics[width=0.98\linewidth]{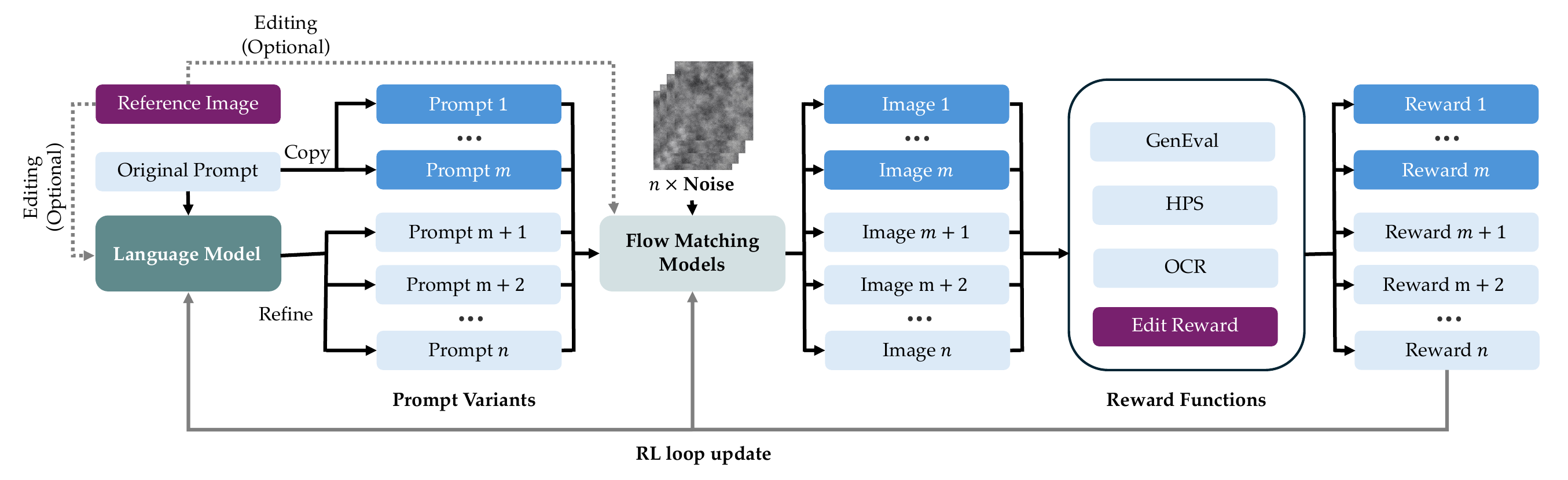}
    \caption{\textbf{Overview of the PromptRL framework.}
    PromptRL jointly trains a language model and a flow-matching image generator within a unified RL loop. Given an original prompt (and optionally a reference image), the LM produces semantically grounded prompt variants that expand the exploration space beyond fixed-prompt training. These prompts are paired with independent noise samples and passed to the flow-matching model to generate diverse images. A mixture of reward functions evaluates each image and guides the evolution of the LM~(for improved prompt rewriting) and the FM~(for improved visual generation). }
    \label{fig:main}
\end{figure*}

The advent of flow matching models (FMs)~\cite{liu2022flow,lipman2022flow,flux2024,song2020denoising} has transformed text-to-image (T2I) generation, enabling photorealistic synthesis from natural language descriptions. To align these models with human preferences and specific reward objectives, reinforcement learning (RL)~\cite{sutton2018reinforcement,fan2023reinforcement,black2023training} has become the standard post-training mechanism, refining model behavior beyond the scope of supervised pretraining. Despite these advances, applying RL to FMs remains prohibitively sample-inefficient. 

Our investigation reveals two underappreciated yet critical failure modes in current flow-based RL pipelines. (\emph{i}) First, we observe a counterintuitive \textit{exploration paradox}: as T2I models improve at following prompts precisely, they simultaneously lose generative diversity under identical prompts. This increased prompt adherence constrains the behavioral variation necessary for effective RL exploration, causing optimization to stagnate in narrow modes of the generation space. (\emph{ii}) Second, we identify severe \textit{prompt overfitting}, where models learn to exploit superficial linguistic patterns in training prompts rather than developing genuine visual understanding. This overfitting manifests as dramatic performance collapse when models encounter semantically equivalent prompts phrased with different syntax at test time. This makes prompt enhancement (PE)~\citep{hao2023optimizing,rosenman2024neuroprompts,manas2024improving,mo2024dynamic}, a crucial technique for improving generation quality, ineffective or even counterproductive for RL-finetuned FMs. We provide detailed empirical evidence for both phenomena in \cref{sec:und}.

These limitations expose a fundamental design oversight in existing approaches: treating prompts as fixed inputs rather than malleable components of the optimization process. Naive augmentation techniques such as random synonym substitution or rule-based paraphrasing prove inadequate, often failing to generate semantically coherent variations at scale. In this paper, we explore the hypothesis that large language models (LMs), when trained as adaptive co-learners via joint RL, can generate semantically grounded prompt variations that enhance exploration efficiency and serve as a co-trained  PE module in practical deployment.

We introduce \textbf{PromptRL} (\textbf{P}rompt \textbf{M}atters in \textbf{RL}), a framework that integrates LMs as adaptive co-learners within flow-based RL training loops, as illustrated in \cref{fig:main}. Rather than employing LMs as static preprocessors, we train them to generate prompt variations that simultaneously preserve semantic intent and maximize downstream image generation rewards. This creates a mutually beneficial training dynamic: diverse LM-generated prompts expand the exploration space for FMs, accelerating policy improvement, while reward signals from flow model outputs guide LMs toward discovering linguistically varied yet contextually appropriate reformulations. 

Our experimental results demonstrate that PromptRL achieves state-of-the-art performance across multiple benchmarks, obtaining scores of 0.97 on GenEval~\citep{ghosh2023geneval}, 0.98 on OCR accuracy~\citep{cui2025paddleocr}, and 24.05 on PickScore~\citep{kirstain2023pick}. Furthermore, we validate the effectiveness of our RL approach on large-scale image editing models, improving EditReward~\citep{wu2025editreward} of FLUX.1-Kontext~\citep{labs2025flux} from 1.19 to 1.43 with only 0.06 million rollouts. Extensive experiments show that PromptRL consistently achieves higher performance ceilings while requiring up to $2\times$ fewer rollouts compared to existing methods, all while maintaining robust generalization to diverse prompt formulations. These findings establish language-vision co-optimization as a foundational principle for efficient and robust preference learning in generative models.

\begin{figure*}[t]
  \centering
  \setlength\tabcolsep{2pt}
  \begin{tabular}{>{\centering\arraybackslash}m{0.2\textwidth} 
                  >{\centering\arraybackslash}m{0.22\textwidth} 
                  >{\centering\arraybackslash}m{0.22\textwidth} 
                  >{\centering\arraybackslash}m{0.22\textwidth}}
    \toprule
    \textbf{Prompt} & \textbf{(a) Stable Diffusion v1-5} & \textbf{(b) FLUX.1-dev} & \textbf{(c) FLUX.1-dev + LM refinement} \\
    \midrule[0.8pt]

    \raggedright\small \textit{Mini bee with nebula wings harvesting honey inside a floating dewdrop kingdom.}
    &
    \begin{subfigure}[b]{\linewidth}
      \centering
      \includegraphics[width=\textwidth]{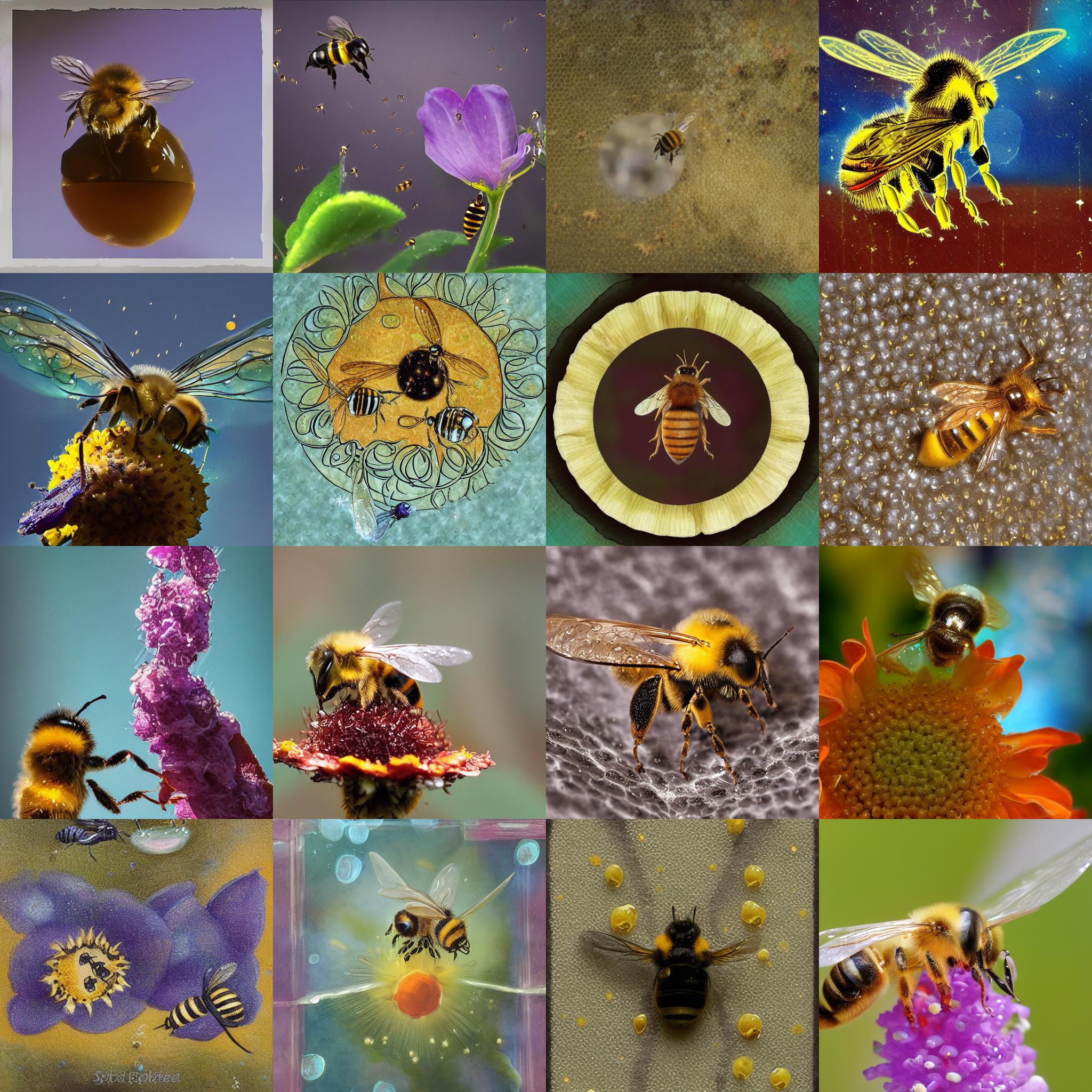}
      \caption*{\scriptsize TI-Sim 0.28$\uparrow$, P.S. 19.4$\uparrow$, II-Sim 0.72$\downarrow$}
      \label{fig:sd15bee}
    \end{subfigure}
    &
    \begin{subfigure}[b]{\linewidth}
      \centering
      \includegraphics[width=\textwidth]{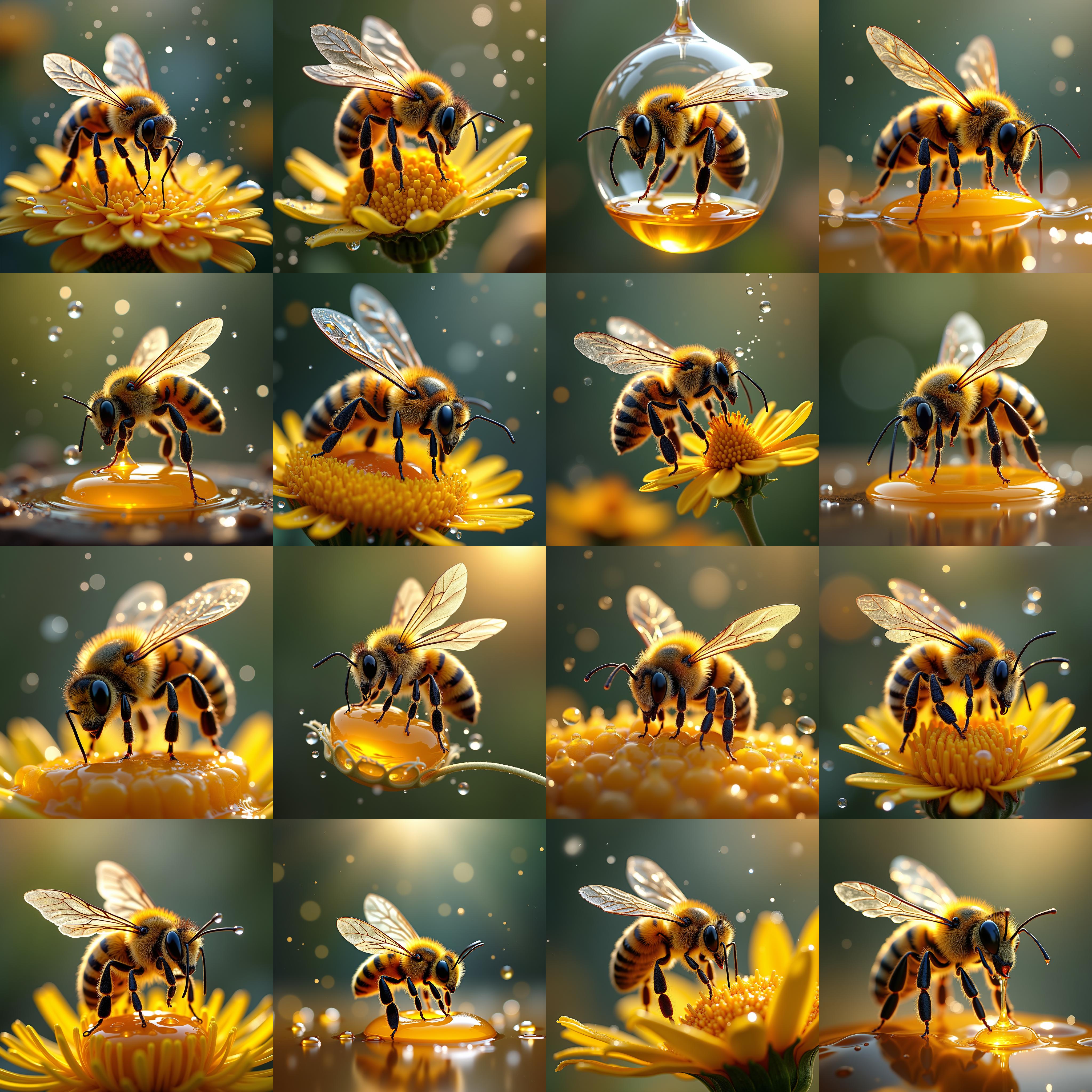}
      \caption*{\scriptsize TI-Sim 0.32$\uparrow$, P.S. 21.9$\uparrow$ II-Sim 0.93$\downarrow$}
      \label{fig:fluxbee}
    \end{subfigure}
    &
    \begin{subfigure}[b]{\linewidth}
      \centering
      \includegraphics[width=\textwidth]{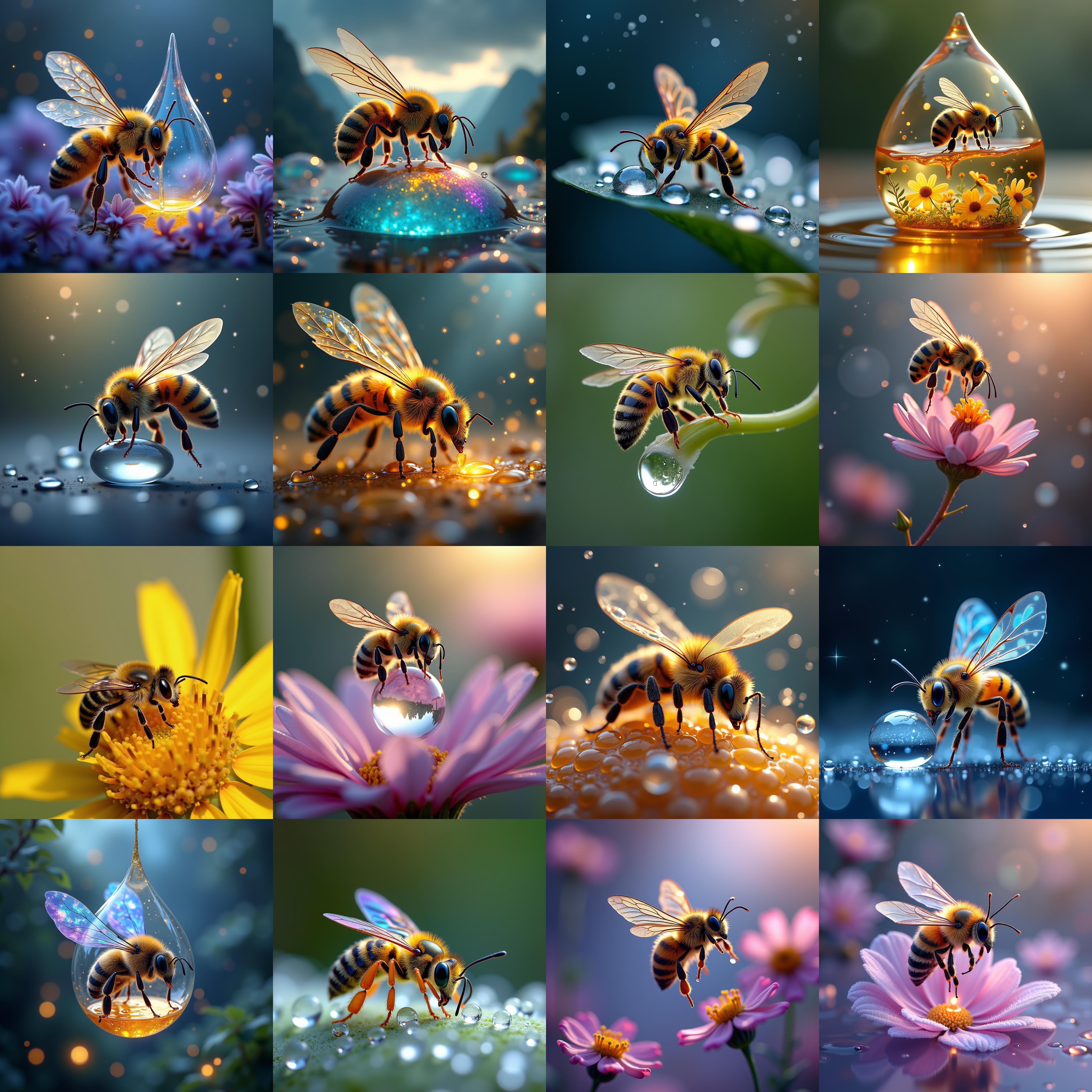}
      \caption*{\scriptsize TI-Sim 0.34$\uparrow$, P.S. 22.3$\uparrow$, II-Sim 0.85$\downarrow$}
      \label{fig:cotbee}
    \end{subfigure}
    \\[12pt]

    \raggedright\small \textit{A tiny koala in liquid mercury spacesuit riding a singing peppermint comet.}
    &
    \begin{subfigure}[b]{\linewidth}
      \centering
      \includegraphics[width=\textwidth]{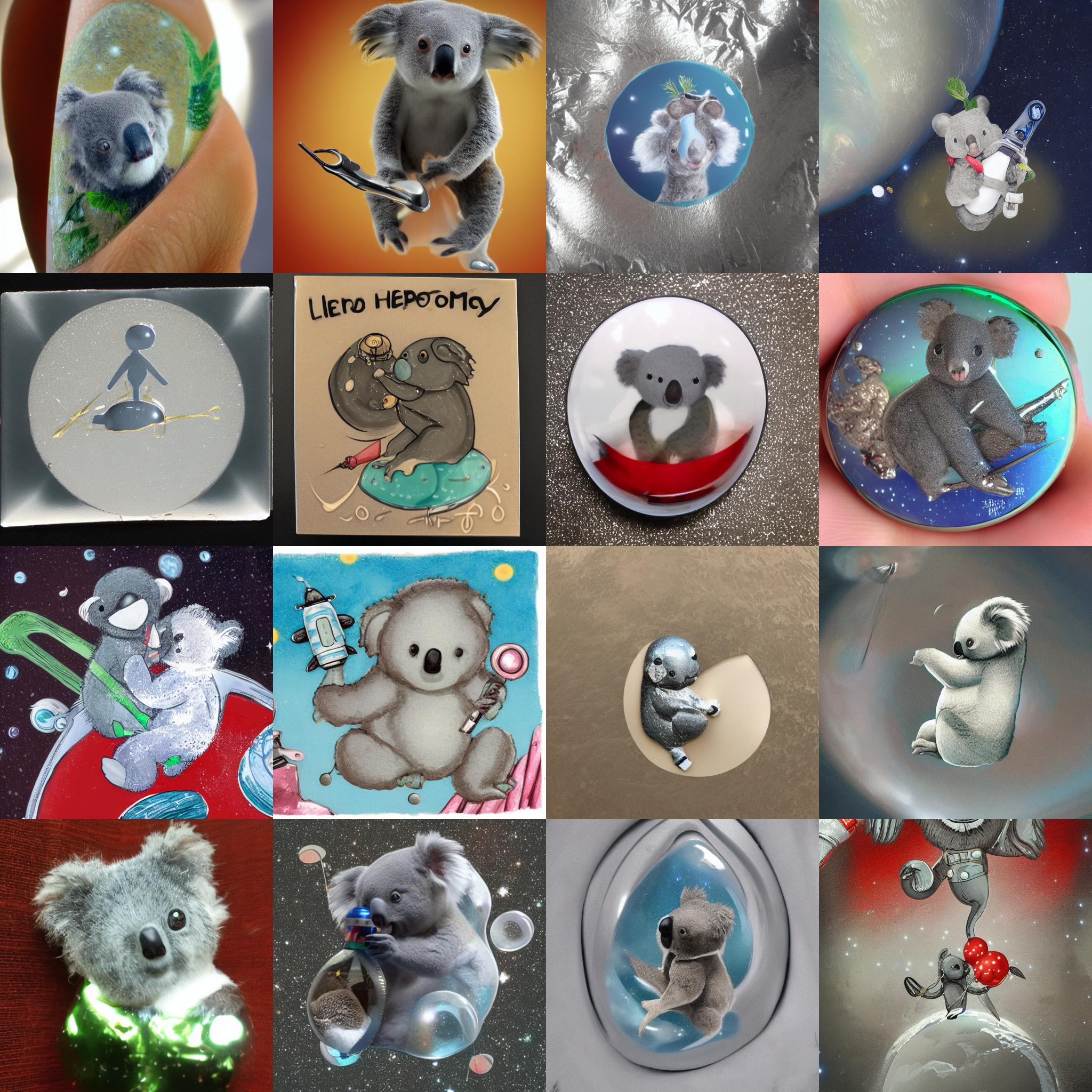}
      \caption*{\scriptsize TI-Sim 0.29$\uparrow$, P.S. 18.9$\uparrow$, II-Sim 0.58$\downarrow$}
      \label{fig:sd15koala}
    \end{subfigure}
    &
    \begin{subfigure}[b]{\linewidth}
      \centering
      \includegraphics[width=\textwidth]{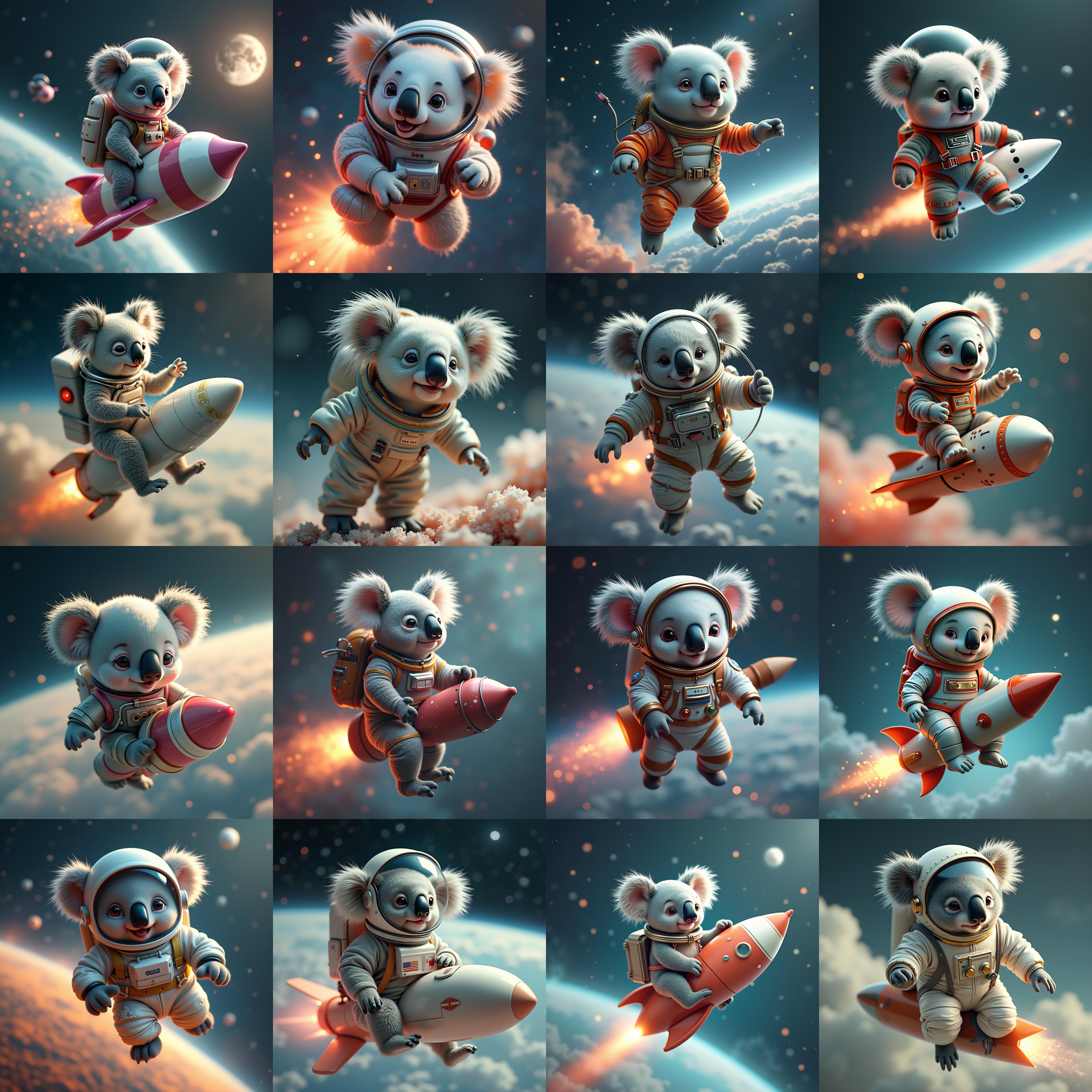}
      \caption*{\scriptsize TI-Sim 0.35$\uparrow$, P.S. 23.0$\uparrow$ II-Sim 0.92$\downarrow$}
      \label{fig:fluxkoala}
    \end{subfigure}
    &
    \begin{subfigure}[b]{\linewidth}
      \centering
      \includegraphics[width=\textwidth]{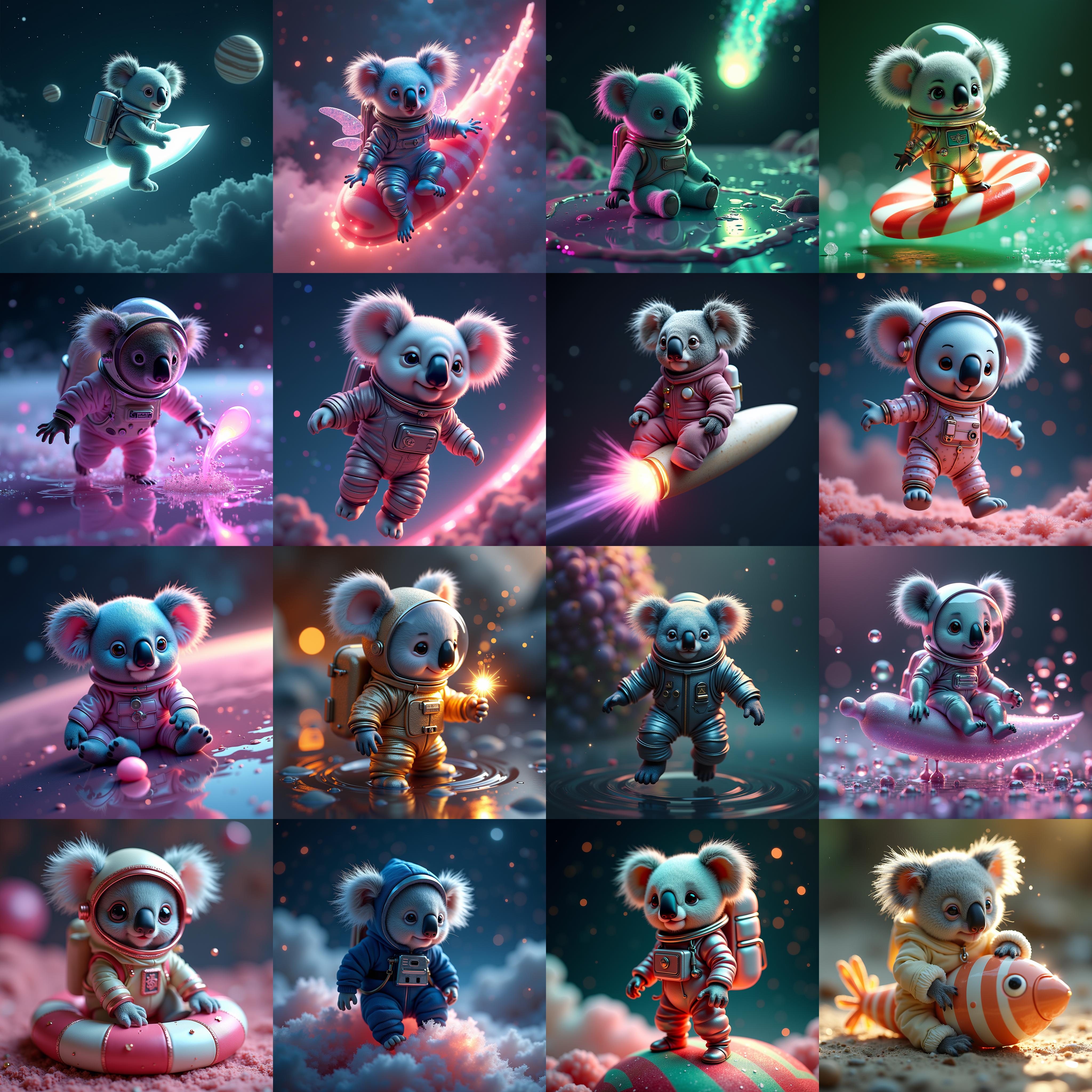}
      \caption*{\scriptsize TI-Sim 0.36$\uparrow$, P.S. 22.9$\uparrow$, II-Sim 0.84$\downarrow$}
      \label{fig:cotkoala}
    \end{subfigure}
    \\
    
    \bottomrule
  \end{tabular}
  \caption{\textbf{The quality-diversity dilemma in flow-based T2I models and its mitigation through prompt refinement.}
As models advance from Stable Diffusion v1-5~(a) to FLUX.1-dev~(b), they achieve higher text-image alignment (TI-Sim) and aesthetic quality (P.S.) but suffer from dramatically reduced output diversity (II-Sim), creating an exploration bottleneck for RL optimization. LM-based prompt refinement~(c) partially restores diversity while maintaining quality, demonstrating that linguistic variations can expand the exploration space. All images in each row share identical random seeds to isolate the effect of  prompt conditioning.}
  \label{fig:qualityvsdiversity}
  \vspace{-4mm}
\end{figure*}

\section{Related works}
\label{sec:related_works}

\noindent \textbf{RL for image generation.}
Reinforcement learning for flow-based image generation has evolved through several paradigms. Early differentiable reward methods (\eg, DRaFT~\citep{clark2023directly}, AlignProp~\citep{prabhudesai2023aligning}, ReFL~\citep{xu2023imagereward}) backpropagate gradients from pre-trained reward models like ImageReward, offering simplicity but prone to reward hacking such as oversaturation. RL-based approaches (\eg, DDPO~\citep{black2023training}, DPOK~\citep{fan2023reinforcement}) treat denoising as an MDP and apply PPO or variants for alignment, with scaled versions showing promise~\citep{zhang2024large}. Direct preference optimization advanced with Diffusion-DPO~\citep{wallace2024diffusion} for simulation-free training on preference pairs, later extended by D3PO~\citep{yang2024using} and SPO~\citep{liang2024step} to combine RL and direct objectives without absolute scores. DiffusionNPO~\citep{wang2025diffusion}, which uses negative/reversed preferences to avoid undesirable modes and boost details, lighting, and structure. Flow-GRPO~\citep{liu2025flow}, DanceGRPO~\citep{xue2025dancegrpo}) extends DDPM-based RL approaches~\cite{black2023training} to flow-matching models by converting flow ODEs to SDEs. Complementary works include DiffusionNFT~\citep{zheng2025diffusionnft}, which performs efficient online RL on the forward noising process via flow matching—contrasting positive/negative samples without reverse gradients, trajectory storage, or solver constraints.  MixGRPO~\citep{li2025mixgrpo} further improves GRPO efficiency in flow settings through hybrid ODE–SDE sampling and sliding-window scheduling. 

\noindent \textbf{PE for image generation}
Prompt enhancement (PE) has become essential for improving text-to-image (T2I) generation quality and alignment. While early approaches relied on manual refinement, recent methods leverage LMs for automated prompt optimization. Promptist~\cite{hao2023optimizing} combines supervised fine-tuning with RL to optimize prompts for aesthetic appeal while preserving user intent. NeuroPrompts~\cite{rosenman2024neuroprompts} introduces constrained text decoding for automatic prompt enhancement with user-controllable styles. OPT2I~\cite{manas2024improving} iteratively refines prompts using LMs to maximize consistency scores. RePrompt~\cite{wu2025reprompt} incorporates chain-of-thought reasoning and reward-guided training for structured reprompting. PAE~\cite{mo2024dynamic} dynamically optimizes word weights and injection timesteps via online RL with multi-objective rewards. However, these methods are limited to prompt-level RL with modest gains—for instance, RePrompt only improves FLUX.1-dev's GenEval score from 0.66 to 0.76—and lack exploration of synergistic prompt-level and flow-level RL. Beyond generation, recent works adopt LMs for instruction-based image editing. MGIE~\cite{fu2023guiding} transforms brief user instructions into expressive editing guidance. SmartEdit~\cite{huang2024smartedit} introduces a Bidirectional Interaction Module for complex reasoning scenarios. Emu Edit~\cite{sheynin2024emu} incorporates the LM to output detailed editing instruction and target image caption. 

\section{Understanding flow RL inefficiencies}
\label{sec:und}

\subsection{Generation dilemma: quality \vs diversity}\label{sec:dilemma}

We observe a fundamental tension between generation quality and output diversity in modern T2I FMs. As models advance in their capacity to precisely follow textual prompts, they simultaneously sacrifice the generative variability essential for effective RL exploration. To quantify this phenomenon, we measure text-image similarity (TI-Sim) using CLIP ViT-g-14~\citep{clip} for prompt alignment, PickScore (P.S.) for aesthetic quality, and image-image similarity (II-Sim) via CLIP ViT-g-14 for output diversity under identical prompts. \cref{fig:qualityvsdiversity} demonstrates this quality-diversity trade-off: Stable Diffusion v1-5 (column a) produces generations with moderate TI-Sim and P.S., but maintains substantial diversity (II-Sim of 0.58--0.72). In contrast, FLUX.1-dev (column b) achieves notably higher aesthetic scores, yet generates outputs with considerably reduced diversity (II-Sim of 0.92--0.93). This pattern holds consistently across semantically diverse prompts, from fantastical creatures to surreal compositional scenes. Notably, LM-based prompt refinement (column c) partially mitigates this collapse, reducing II-Sim to 0.84--0.85 while preserving quality metrics. 

This quality-diversity dilemma directly undermines RL optimization in flow-based models. As generation policies become increasingly deterministic, rollout trajectories collapse into narrow modes of the output space, causing reward signals to degenerate: when all samples cluster around similar high-quality outputs, advantage estimators lose the comparative information necessary for policy improvement. This exploration bottleneck is further compounded by severe prompt overfitting, wherein models exploit superficial lexical patterns rather than semantic understanding, which we investigate in the following subsection.

\subsection{Prompt linguistic hacking}

\begin{table}[t]
\centering
\caption{Performance comparison across different models and evaluation metrics. All metrics are tested with 20 Euler steps with resolution 1024. PE denotes prompt enhancement with Qwen-2.5-VL. \textcolor{green}{Green} and \textcolor{red}{red} numbers indicate performance changes after applying PE. P.S. and U.R. denotes PickScore and UnifiedReward, respectively.} 
\label{tab:collapse}
\resizebox{\columnwidth}{!}{%
\begin{tabular}{llccccc}
\toprule
\textbf{Model} & \textbf{w/ PE} & \textbf{GenEval} & \textbf{OCR} & \textbf{P.S.} & \textbf{HPS} & \textbf{U.R.} \\
\midrule
SD3 & \xmark & 0.58 & 0.48 & 22.30 & 26.95 & 2.982 \\
SD3 & \cmark & 0.63 \textcolor{green}{\scriptsize(+0.05)} & 0.53 \textcolor{green}{\scriptsize(+0.05)} & 22.34 \textcolor{green}{\scriptsize(+0.04)} & 27.67 \textcolor{green}{\scriptsize(+0.72)} & 3.140 \textcolor{green}{\scriptsize(+0.158)} \\
\midrule
DiffusionNFT & \xmark & 0.88 & 0.89 & 23.63 & 31.78 & 3.392 \\
DiffusionNFT & \cmark & 0.77 \textcolor{red}{\scriptsize(-0.11)} & 0.86 \textcolor{red}{\scriptsize(-0.03)} & 23.21 \textcolor{red}{\scriptsize(-0.42)} & 30.66 \textcolor{red}{\scriptsize(-1.12)} & 3.268 \textcolor{red}{\scriptsize(-0.124)} \\
\midrule
FlowGRPO & \xmark & 0.92 & 0.89 & 23.33 & — & — \\
FlowGRPO & \cmark & 0.81 \textcolor{red}{\scriptsize(-0.11)} & 0.86 \textcolor{red}{\scriptsize(-0.03)} & 23.13 \textcolor{red}{\scriptsize(-0.20)} & — & — \\
\bottomrule
\end{tabular}%
}
\end{table}

Beyond the quality-diversity dilemma, we identify a second critical failure mode: \textit{prompt linguistic hacking}, where RL-trained models exploit superficial lexical patterns rather than developing robust semantic understanding. We evaluate this by testing models on both original prompts and semantically-preserved paraphrases generated by Qwen-2.5-VL~(\ie, PE)~\citep{bai2025qwen2}.

As shown in \cref{tab:collapse}, the pretrained SD3~\citep{esser2024scaling} demonstrates linguistic robustness, with consistent or improved performance under paraphrasing across all metrics. However, flow-only RL models exhibit severe prompt overfitting. DiffusionNFT achieves strong performance on original prompts but suffers catastrophic degradation under paraphrasing. Similarly, FlowGRPO trained on GenEval drops from 0.92 to 0.81 when prompts are paraphrased. This indicates that \textit{learned policies memorize superficial linguistic features} rather than understanding underlying visual concepts. This overfitting occurs at the \textit{prompt-conditioning} level and cannot be resolved through standard regularization. More critically, \textit{PE techniques that benefit pretrained FMs become ineffective or even harmful after flow-only RL}, as fine-tuned models overfit to specific prompt distributions. This observation motivates our joint LM-FM optimization approach: rather than applying PE as a fixed preprocessing step, we co-evolve the prompt enhancer and generator in a symbiotic manner.

\section{PromptRL}
\label{sec:PromptRL}
\begin{figure*}[t]
    \centering
    \includegraphics[width=0.95\linewidth]{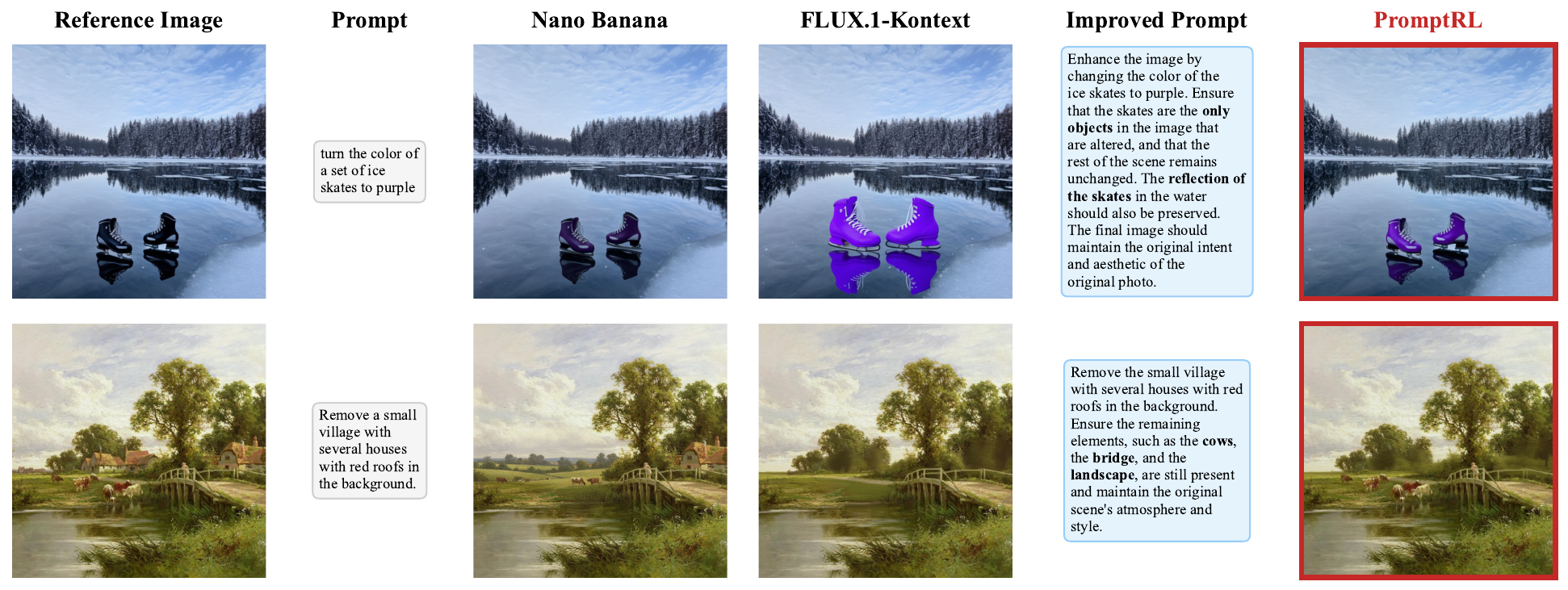}
    \caption{Qualitative comparison on instructional image editing tasks. 
Our method enables the LM to leverage the original image's visual signals to transform vague editing instructions into more explicit and image-specific prompts, ultimately improving editing performance.
    }
    \label{fig:edit_comparison}
\end{figure*}

\subsection{Incorporating LMs as dynamic prompt refiner}
\label{sec:lm_refiner}

To address the dual challenges of exploration collapse and prompt overfitting identified in \cref{sec:und}, we propose incorporating LMs as adaptive prompt refiners within the RL training loop. Unlike static augmentation techniques that rely on rule-based transformations or synonym substitution, our approach leverages the semantic understanding and compositional flexibility of pretrained LMs to generate contextually grounded prompt variations.

Formally, given an original prompt $p_0$ from the training distribution, we deploy an LM $\pi_{\text{LM}}(\cdot | p_0)$ to generate a set of refined prompts $\{p_1, p_2, \ldots, p_k\}$ that preserve the core semantic intent while introducing linguistic diversity. Each refined prompt $p_i$ is then paired with an independent noise sample $\boldsymbol \epsilon_i \sim \mathcal{N}(0, I)$ and fed into the flow-matching model $\pi_{\text{FM}}(\cdot | p_i, \epsilon_i)$ to produce diverse image samples $\{\mathbf x_1, \mathbf x_2, \ldots, \mathbf x_k\}$. This architecture creates a hierarchical exploration mechanism: the LM explores the linguistic manifold of semantically equivalent descriptions, while the FM explores the visual manifold conditioned on each prompt variant.

Critically, we introduce a \textit{prompt retention mechanism} during training: for each batch of $n$ total samples, we retain $m < n$ samples that use the original prompt $p_0$ without LM refinement, while the remaining $n - m$ samples undergo LM-based augmentation. This design serves two complementary purposes. First, unmodified prompts provide a \textit{strong baseline} for advantage estimation—augmented prompts that yield lower rewards than $p_0$ are effectively pruned during policy optimization, preventing wasteful exploration in low-reward regions of the prompt space. Second, consistent exposure to original prompts ensures that the FM maintains robust performance on the training distribution, preventing the model from becoming overly dependent on LM refinements at inference time.

\subsection{Joint RL training on disjoint LMs and FMs}
\label{sec:joint_training}

Having established the LM as a dynamic prompt refiner, we now describe the joint RL training procedure that simultaneously optimizes both $\pi_{\text{LM}}$ and $\pi_{\text{FM}}$ within a unified policy gradient framework. Crucially, while the two models share reward signals, they remain architecturally \textit{disjoint}—gradients do not propagate between the LM and FM, preserving modularity and computational efficiency.

At each training iteration, we sample a batch of $B$ original prompts $\{p_0^{(1)}, \ldots, p_0^{(B)}\}$ from the training distribution. For each prompt $p_0^{(j)}$, we generate $n$ samples using the procedure described in \cref{sec:lm_refiner}: $m$ samples are generated directly from $p_0^{(j)}$ with different noise seeds, while the remaining $n - m$ samples use LM-refined prompts $\{p_i^{(j)}\}_{i=1}^{n-m}$ sampled from $\pi_{\text{LM}}(\cdot | p_0^{(j)})$. Each prompt (original or refined) is paired with an independent latent noise vector $\boldsymbol \epsilon \sim \mathcal{N}(0, I)$ to produce images via the FM.

\begin{figure*}[t]
    \centering
    \includegraphics[width=0.95\linewidth]{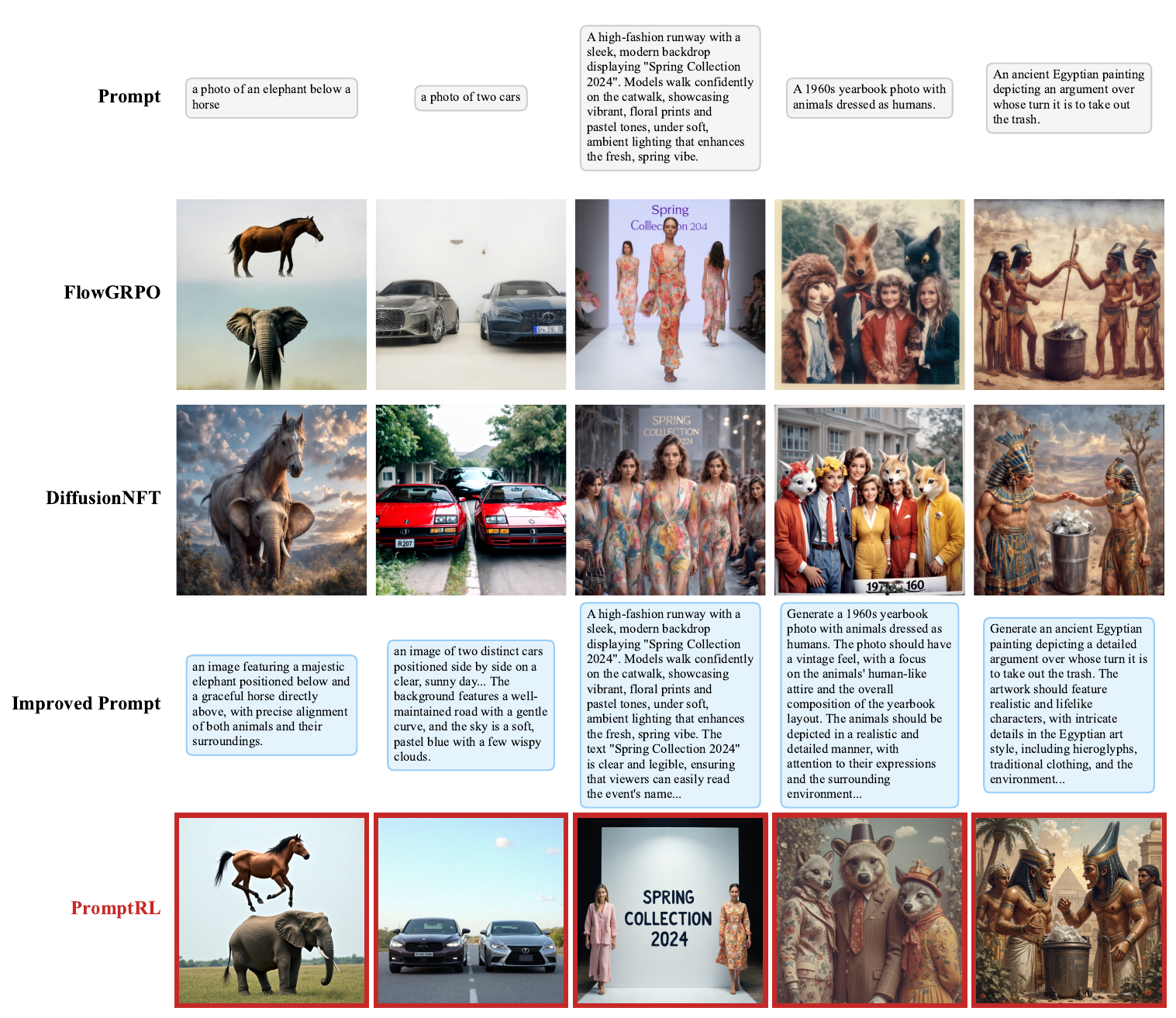}
    \caption{Qualitative comparison on text-to-image generation. The first two prompts are from GenEval. The third prompt is from OCR-1k. And the last two prompts are from Drawbench.
    }
    \label{fig:t2i_comparison}
\end{figure*}
The resulting images are evaluated using a composite reward function $R(\cdot)$ that aggregates format reward and image generation reward:
\begin{equation}
R(\mathbf x, p) = \lambda_{\text{Format}} R_{\text{Format}}(p) + \lambda_{\text{Gen}} R_{\text{Gen}}(\mathbf x, p),
\end{equation}
where $R_{\text{Format}}(p)$ is a binary reward that requires the LM to output refined prompt variants enclosed within XML tags \texttt{<answer>} and \texttt{</answer>}. This format constraint ensures structured output parsing and penalizes malformed generations, with $R_{\text{Format}}(p) = 1$ if the output conforms to the required format and $0$ otherwise. The term $R_{\text{Gen}}(\mathbf x, p)$ denotes a image generation reward models (\eg, GenEval for compositional accuracy). The hyperparameters $\lambda_{\text{Format}}$ and $\lambda_{\text{Gen}}$ balance format compliance and generation quality; in practice, we set $\lambda_{\text{Format}} = 1.0$ and $\lambda_{\text{Gen}} = 1.0$ for simplicity. Following the GRPO, we compute advantages through group-wise normalization. For each prompt $p_0^{(j)}$ and its $n$ associated samples $\{\mathbf x_1^{(j)}, \ldots, \mathbf x_n^{(j)}\}$, we calculate:
\begin{equation}
A(\mathbf x_i^{(j)}, p_i^{(j)}) = \frac{R(\mathbf x_i^{(j)}, p_i^{(j)}) - \mu^j}{\sigma^j + \epsilon},
\end{equation}
where $\mu^j$ and $\sigma^j$ are the mean and standard deviation of rewards within the $j$-th group, and $\epsilon$ is a small constant for numerical stability. This group-wise normalization makes advantage signals invariant to reward scale across different prompts, enabling stable optimization across diverse semantic categories. It also inherently implements a form of self-competition where samples generated from the same prompt compete against each other, encouraging within-group diversity.

Crucially, while the LM and FM share reward signals in PromptRL, they remain architecturally \textit{disjoint}—gradients do not propagate between the LM and FM, preserving modularity and computational efficiency. For the language model, we only retain advantages corresponding to the $n - m$ LM-refined samples, excluding those generated from original prompts. The LM policy gradient is:
\begin{equation}
\nabla_{\theta_{\text{LM}}} \mathcal{J}_{\text{LM}} = \mathbb{E}_{p_0} \left[ \sum_{i=m+1}^{n} A(\mathbf x_i, p_i) \cdot \nabla_{\theta_{\text{LM}}} \log \pi_{\text{LM}}(p_i | p_0) \right].
\end{equation}
This selective advantage assignment ensures the LM learns to generate prompt variants that improve upon the baseline. Variants that underperform receive negative advantages and are down-weighted, preventing the LM from introducing counterproductive linguistic changes.

For the flow matching model, we utilize advantages from all $n$ samples (both original and refined prompts) to update the image generation policy:
\begin{equation}
\nabla_{\theta_{\text{FM}}} \mathcal{J}_{\text{FM}} = \mathbb{E}_{p_0} \left[ \sum_{i=1}^{n} A(\mathbf x_i, p_i) \cdot \nabla_{\theta_{\text{FM}}} \log \pi_{\text{FM}}(\mathbf x_i | p_i, \boldsymbol \epsilon_i) \right] \, .
\end{equation}
 The inclusion of advantages from original prompts ensures the FM maintains strong performance on the base distribution while benefiting from the expanded exploration space provided by LM refinements.

Notably, this joint optimization requires no architectural/algorithmic modifications to either model—we simply backpropagate separate policy gradients through their respective parameters using shared rewards as the update signal. Our framework is agnostic to the specific RL algorithm; while we use GRPO~\citep{shao2024deepseekmath} as the baseline implementation for its simplicity, the components of PromptRL should be able to integrated with other online RL approaches for LMs~(\eg, ReMax~\citep{li2023remax}) and FMs~(\eg, DiffusionNFT~\cite{zheng2025diffusionnft}).

\subsection{Multi-reward training via reward tagging} 
Beyond single-reward scenarios, we validate PromptRL under multi-reward optimization with GenEval, PickScore, and OCR. A known challenge is that different reward models exhibit vastly different scales and variances, requiring cumbersome manual tuning of reward coefficients. Alternative approaches adopt multi-stage training pipelines (\eg, DiffusionNFT first trains on PickScore before transitioning to additional rewards), introducing scheduling complexity.
We propose a simple yet effective solution: \textit{single-reward-per-sample training}. Rather than computing weighted reward sums, we assign each training prompt a categorical tag indicating which reward function evaluates its generated images. During group-wise advantage computation, normalization is performed within each reward category, ensuring comparable advantage signals despite differing scales. This eliminates reward coefficient tuning entirely—each reward model operates in its native scale without interference. Empirically, this simple tagging mechanism achieves strong performance across all objectives simultaneously without explicit reward engineering or multi-stage curricula.

\section{Experiments}

\subsection{Experimental setup}

\noindent \textbf{Base models.} To comprehensively validate PromptRL on flow-based image generation, we evaluate on both T2I synthesis and instructional image editing. We adopt FLUX.1-dev as the base flow-matching model for T2I generation and FLUX.1-Kontext for image editing. For the language model component, we use Qwen2.5-VL-3B-Instruct as the prompt refiner.

\noindent \textbf{Datasets and benchmarks.} For text-to-image generation, following FlowGRPO, we evaluate across three complementary dimensions: compositional accuracy (GenEval), text rendering capability (OCR), and human preference alignment (PickScore). For GenEval and OCR objectives, we use the corresponding training sets from FlowGRPO. For PickScore optimization, we train on the Pick-a-Pic dataset. For instructional image editing, we randomly sample 10,000 examples from the OmniEdit~\citep{wei2024omniedit} training set, retaining only the editing instructions and corresponding reference images. For T2I generation, we use the official GenEval validation set for compositional evaluation, DrawBench~\citep{drawbench} for aesthetic quality assessment (PickScore, HPS~\citep{wu2023hpsv2}, UnifiedReward~\citep{wang2025unified}), and three OCR benchmarks: the FlowGRPO OCR-1k validation set, TMDB, and OpenLib from MARIO-Eval~\citep{chen2023textdiffuser}. For image editing, we directly evaluate on the OmniEdit validation set, which covers six editing categories: Object Swap, Object Addition, Object Removal, Attribute Modification, Environment Change, and Style Transfer.

\noindent \textbf{Training details.} For computational efficiency, we conduct rollouts at 512$\times$512 resolution with 20 inference steps when training FLUX.1-dev. However, we observe that reducing resolution significantly degrades the editing capabilities of FLUX.1-Kontext; therefore, we maintain 1024$\times$1024 resolution for editing experiments. To improve training efficiency under the high computational demands of high-resolution training, we reduce the number of inference steps to 8 and apply SDE solving only during the first 4 denoising steps of each rollout, following prior work~\citep{li2025mixgrpo} demonstrating that image structure largely stabilizes in the early timesteps. All models are trained using the GRPO algorithm with a group size of $n=8$ samples per prompt and a prompt retention number of $m=2$.

\subsection{Quantitative comparison}
\noindent \textbf{Text-to-image generation.} As shown in \cref{tab:geneval}, PromptRL achieves state-of-the-art performance on GenEval with an overall score of \textbf{0.97}, outperforming FlowGRPO at 0.92 and DiffusionNFT at 0.88. PromptRL w/ PE attains near-perfect scores of 0.99 on both Position and Counting, demonstrating exceptional compositional accuracy. Notably, even without prompt enhancement at inference, PromptRL w/o PE achieves 0.94, confirming that joint training instills robust visual understanding independent of LM refinements.
\cref{tab:performance} further validates PromptRL across aesthetic and OCR metrics. Our method achieves a PickScore of 24.04, HPS of 32.03, and OCR accuracy of 0.98 on OCR-1k, consistently surpassing prior RL-based approaches. These improvements across diverse metrics demonstrate that joint LM-FM optimization generalizes beyond single-objective reward optimization.

\noindent \textbf{Instructional image editing.} For instructional image editing, as reported in \cref{tab:edit_performance}, PromptRL w/ PE improves upon the FLUX.1-Kontext baseline from 1.19 to \textbf{1.43}, approaching ReasonEdit-Think at 1.44, which relied
on fine-grained and dense data annotations along
with a complex multi-stage training pipeline. Improvements are most pronounced on Removal (+0.69) and Environment (+0.28). Notably, naively applying prompt enhancement without joint training degrades FLUX.1-Kontext performance from 1.19 to 1.01, demonstrating that PromptRL's joint optimization is essential for effective prompt refinement.

\begin{table}[t]
\centering
\caption{Performance comparison on GenEval benchmark across different models. Higher scores indicate better performance. Best results are shown in \textbf{bold}. Metrics of models with * are obtained from the Qwen-Image paper.}
\label{tab:geneval}
\resizebox{\linewidth}{!}{%
\begin{tabular}{lcccccc|c}
\toprule
\textbf{Model} & \textbf{1 Obj.} & \textbf{2 Obj.} & \textbf{Cnt.} & \textbf{Clr.} & \textbf{Pos.} & \textbf{Attr.} & \textbf{Avg.}$\uparrow$ \\
\midrule
Show-o*~\citep{xie2024show} & 0.95 & 0.52 & 0.49 & 0.82 & 0.11 & 0.28 & 0.53 \\
Emu3-Gen*~\citep{wang2024emu3} & 0.98 & 0.71 & 0.34 & 0.81 & 0.17 & 0.21 & 0.54 \\
SD3 Medium*~\citep{esser2024scaling} & 0.98 & 0.74 & 0.63 & 0.67 & 0.34 & 0.36 & 0.62 \\
FLUX.1-dev*~\citep{flux2024} & 0.98 & 0.81 & 0.74 & 0.79 & 0.22 & 0.45 & 0.66 \\
SD3.5 Large* & 0.98 & 0.89 & 0.73 & 0.83 & 0.34 & 0.47 & 0.71 \\
JanusFlow*~\citep{ma2025janusflow} & 0.97 & 0.59 & 0.45 & 0.83 & 0.53 & 0.42 & 0.63 \\
Janus-Pro-7B*~\citep{chen2025janus} & 0.99 & 0.89 & 0.59 & 0.90 & 0.79 & 0.66 & 0.80 \\
HiDream~\citep{cai2025hidream} & 1.00 & 0.98 & 0.79 & 0.91 & 0.60 & 0.72 & 0.83 \\
Seedream 3.0*~\citep{gao2025seedream} & 0.99 & 0.96 & 0.91 & 0.93 & 0.47 & 0.80 & 0.84 \\
Qwen-Image*~\citep{wu2025qwen} & 0.99 & 0.92 & 0.89 & 0.88 & 0.76 & 0.77 & 0.87 \\
\midrule
\multicolumn{8}{l}{\textit{RL-based}} \\
\midrule
\rowcolor{black!5} RePrompt~\citep{wu2025reprompt} & 0.98 & 0.87 & 0.77 & 0.85 & 0.62 & 0.49 & 0.76 \\

\rowcolor{black!5} FlowGRPO~\citep{liu2025flow} & 1.00 & 0.99 & 0.91 & 0.89 & 0.95 & 0.80 & 0.92 \\
\rowcolor{black!5} DiffusionNFT~\citep{zheng2025diffusionnft} & 1.00 & 0.98 & 0.74 & 0.92 & 0.85 & 0.80 & 0.88 \\
\rowcolor{black!5} PromptRL w/o PE & 1.00 & 0.96 & 0.95 & 0.95 & 0.93 & 0.85 & 0.94 \\
\rowcolor{black!5} PromptRL w/ PE & \textbf{1.00} & \textbf{0.99} & \textbf{0.99} & \textbf{0.96} & \textbf{0.99} & \textbf{0.90} & \textbf{0.97} \\
\bottomrule
\end{tabular}%
}
\end{table}

\begin{table}[t]
\centering
\caption{Performance comparison across different models on aesthetic and OCR metrics.}
\label{tab:performance}
\resizebox{0.95\columnwidth}{!}{%
\begin{tabular}{lcccccc}
\toprule
& \multicolumn{3}{c}{\textit{Aesthetic}} & \multicolumn{3}{c}{\textit{OCR}} \\
\cmidrule(lr){2-4} \cmidrule(lr){5-7}
\textbf{Model} & \textbf{P.S.} & \textbf{HPS} & \textbf{U.R.} & \textbf{OCR-1k} & \textbf{TMDB} & \textbf{EOpenLib} \\
\midrule
SD1.5~\citep{rombach2022high} & 20.92 & 23.71 & 2.00 & 0.05 & 0.13 & 0.08 \\
SDXL~\citep{podell2023sdxl} & 22.14 & 26.67 & 2.78 & 0.13 & 0.20 & 0.09 \\
SD3 Medium~\citep{esser2024scaling} & 22.38 & 28.56 & 3.09 & — & 0.44 & 0.33 \\
FLUX.1-schnell~\citep{flux2024} & 22.64 & 29.39 & 3.25 & 0.54 & 0.66 & 0.50 \\
FLUX.2-klein~\citep{flux-2-2025} & 22.79 & 29.03 & 3.29 & 0.55 & 0.22 & 0.46 \\
Z-Image~\citep{cai2025z} & 20.14 & 28.22 & 3.51 & 0.70 & 0.71 & 0.83 \\
Qwen-Image~\citep{wu2025qwen} & 23.05 & 30.40 & 3.53 & 0.65 & 0.79 & 0.94 \\
Qwen-Image-2512 & 23.16 & 30.79 & 3.40 & 0.72 & 0.81 & 0.87 \\
\midrule
\multicolumn{7}{l}{\textit{RL-based}} \\
\midrule
\rowcolor{black!5} FlowGRPO & 23.33 & 29.80 & 3.33 & 0.89 & 0.83 & 0.73 \\
\rowcolor{black!5} DiffusionNFT & 23.63 & 31.79 & 3.39 & 0.89 & 0.91 & 0.86 \\
 \rowcolor{black!5} PromptRL w/o PE & 24.01 & 31.79 & 3.38 & 0.97 & 0.92 & 0.95 \\
\rowcolor{black!5} PromptRL w/ PE & 24.05 & 32.03 & 3.44 & 0.98 & 0.91 & 0.95 \\
\bottomrule
\end{tabular}%
}
\end{table}

\noindent \textbf{Multi-reward training.} We evaluate our tag-based multi-reward training strategy in \cref{tab:multireward}. Despite using no reward coefficient tuning or multi-stage curriculum, the multi-reward model achieves competitive performance across all objectives (GenEval: 0.93, OCR: 0.96, PickScore: 23.94), with only modest degradation compared to single-reward specialists. 
\begin{table}[t]
\centering
\caption{Comparison of single-reward and multi-reward training. Single-reward trains separately on each objective, while multi-reward uses tag-based joint optimization.}
\label{tab:multireward}
\resizebox{0.7\columnwidth}{!}{%
\begin{tabular}{lccc}
\toprule
\textbf{Training} & \textbf{GenEval}$\uparrow$ & \textbf{OCR}$\uparrow$ & \textbf{PickScore}$\uparrow$ \\
\midrule
Single-reward & 0.97 & 0.98 & 24.05 \\
Multi-reward  & 0.93 & 0.96 & 23.94 \\
\bottomrule
\end{tabular}%
}
\end{table}

\begin{table}[t]
\centering
\caption{Performance comparison on image editing tasks measured by EditReward. We evaluate models across six editing categories: Swap, Style, Addition~(Add.), Attribute Modification (attr.), Environment (Env.), and Removal. The editing instructions of  FLUX.1-Kontext w/ PE are refined by pretrained Qwen2.5-VL.}
\label{tab:edit_performance}
\resizebox{\columnwidth}{!}{%
\begin{tabular}{lcccccc|c}
\toprule
\textbf{Model} & \textbf{Swap} & \textbf{Style} & \textbf{Add.} & \textbf{Attr.} & \textbf{Env.} & \textbf{Removal} & \textbf{Avg.}$\uparrow$ \\
\midrule
InstructPix2Pix~\citep{brooks2023instructpix2pix} & -0.24 & 0.91 & -0.45 & 0.45 & 0.48 & -0.80 & 0.02 \\
MagicBrush~\citep{zhang2023magicbrush} & -0.38 & 0.36 & -0.78 & -0.80 & 0.91 & -0.85 & -0.27 \\
LEDITS++~\citep{brack2024ledits++} & -0.81 & -0.32 & -0.30 & -0.60 & -0.37 & -0.97 & -0.60 \\
Qwen-Image-Edit & 1.11 & 1.14 & 0.95 & 0.90 & 1.39 & 0.61 & 1.03 \\
FLUX.2-klein & 1.42 & 1.73 & 1.29 & 1.42 & 1.80 & 0.32 & 1.34 \\
Nano Banana & 1.58 & 1.20 & 1.28 & 1.18 & 1.61 & 1.13 & 1.37 \\
Step1X-Edit~\citep{liu2025step1x} & 1.39 & 1.58 & 1.19 & 1.34 & 1.57 & 0.22 & 1.24 \\
ReasonEdit~\citep{yin2025reasonedit} & 1.51 & 1.43 & 1.19 & 1.47 & 1.58 & 1.14 & 1.40 \\
ReasonEdit-Think & \textbf{1.52} & \textbf{1.47} & 1.19 & \textbf{1.44} & 1.69 & \textbf{1.27} & \textbf{1.44} \\
\rowcolor{black!5} FLUX.1-Kontext & 1.35 & 1.36 & 1.16 & 1.15 & 1.44 & 0.55 & 1.19 \\
\rowcolor{black!5} FLUX.1-Kontext w/ PE & 1.35 & 0.97 & 1.04 & 0.48 & 1.22 & 0.65 & 1.01 \\
\midrule
\rowcolor{black!5} \textbf{PromptRL w/o PE} & 1.45 & 1.46 & 1.28 & 1.35 & 1.56 & 0.98 & 1.36 \\
\rowcolor{black!5} \textbf{PromptRL w/ PE} & \textbf{1.47} & \textbf{1.43} & \textbf{1.29} & \textbf{1.39} & \textbf{1.72} & \textbf{1.24} & \textbf{1.43} \\
\bottomrule
\end{tabular}%
}
\end{table}

\subsection{Qualitative comparison}
\cref{fig:edit_comparison,fig:t2i_comparison} presents qualitative comparisons across T2I generation and image editing tasks. For compositional prompts, PromptRL correctly renders objects with accurate colors and spatial arrangements where baselines exhibit color leakage or omission. On OCR tasks, PromptRL produces legible, correctly spelled text while maintaining aesthetic quality. For instructional editing, PromptRL's jointly-trained LM produces image-aware prompt refinements by leveraging visual signals from reference images, transforming vague instructions into precise, image-specific prompts that preserve foreground subjects while modifying only intended regions.

\subsection{Ablation study}

\begin{figure}[t]
    \centering
    \begin{subfigure}[b]{0.75\linewidth}
        \centering
        \includegraphics[width=\linewidth]{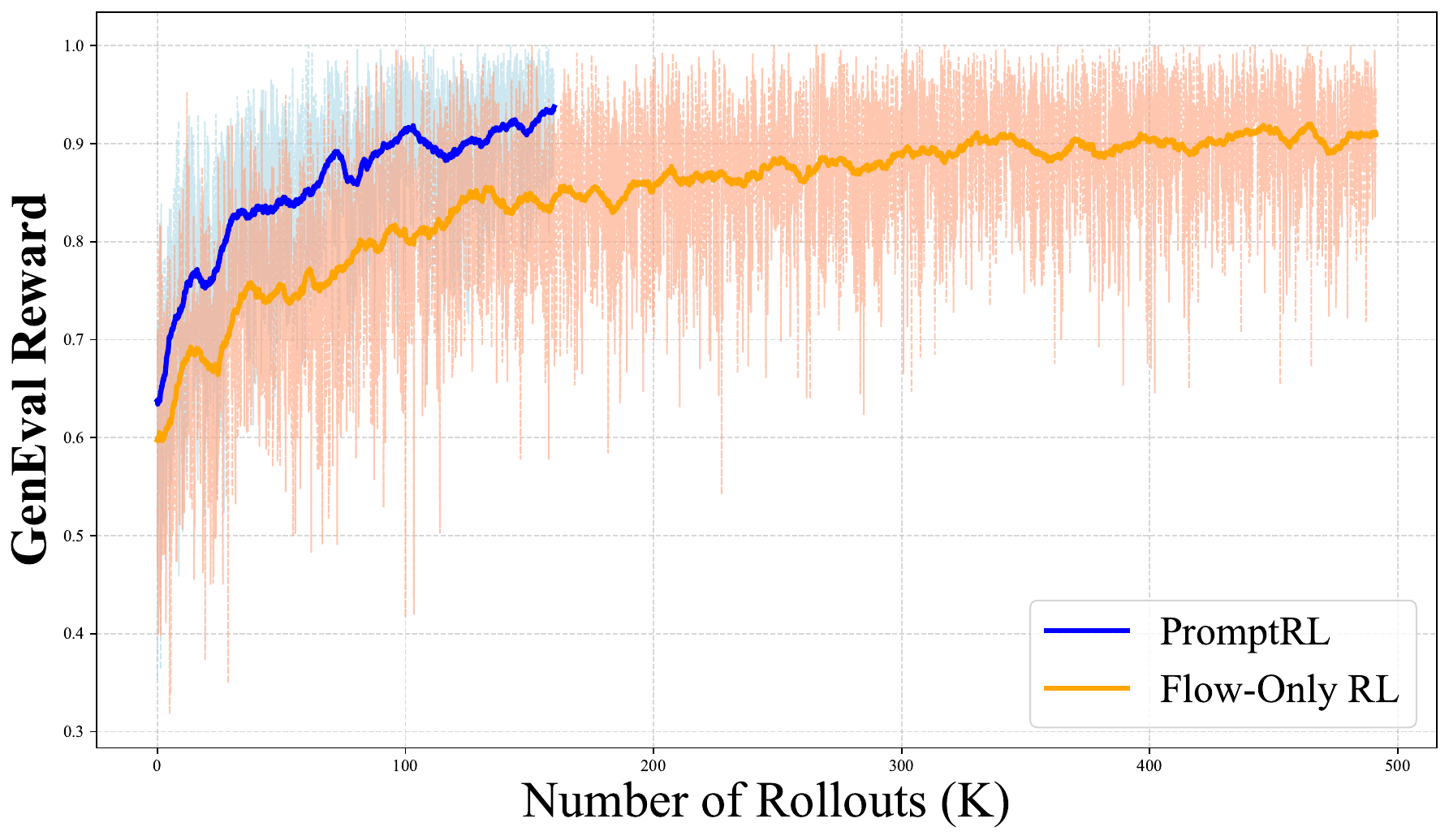}
        \caption{Training curve comparison on GenEval reward.}
        \label{fig:curve_geneval}
    \end{subfigure}    
    \begin{subfigure}[b]{0.75\linewidth}
        \centering
        \includegraphics[width=\linewidth]{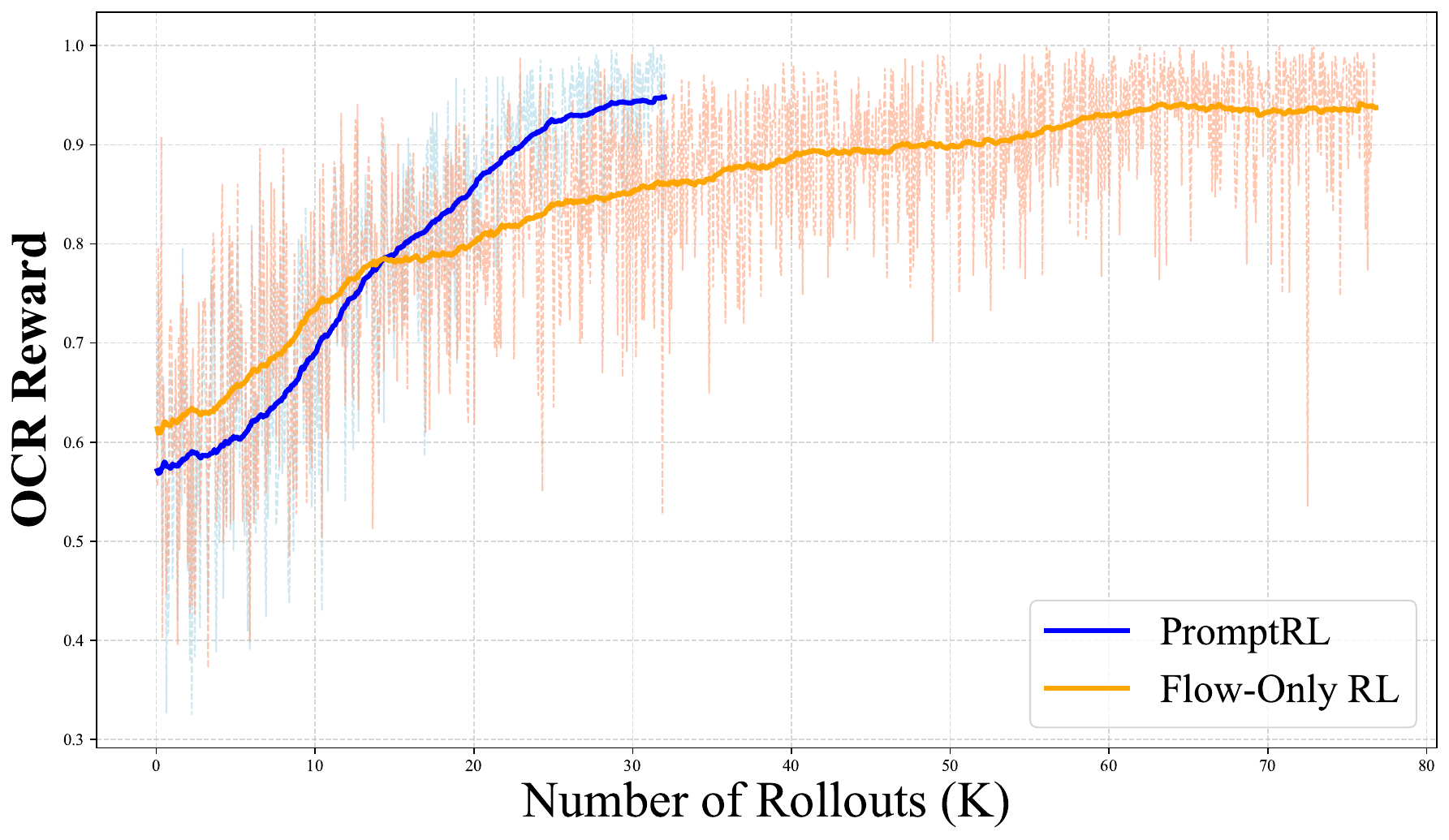}
        \caption{Training curve comparison on OCR reward.}
        \label{fig:curve_ocr}
    \end{subfigure}
    \caption{Training curve comparison on different rewards.}
    \label{fig:training_curves}
\end{figure}

\noindent \textbf{Joint RL boosts training efficiency.}
We compare training dynamics as rollout counts increase, with both methods using FlowGRPO for FM optimization and single-update-per-sample for stability. As shown in \cref{fig:curve_geneval} and \cref{fig:curve_ocr}, PromptRL consistently achieves higher rewards with fewer rollouts across both GenEval and OCR objectives. On GenEval, PromptRL reaches comparable performance to FlowGRPO's convergence point using approximately 50\% fewer rollouts. Similar trends are observed on OCR, where PromptRL demonstrates faster initial improvement and superior final performance. These results validate that the expanded exploration space provided by LM-generated prompt variations improves sample efficiency.

\noindent \textbf{Prompt retention mechanism.}
We investigate the effect of retaining original prompts during training on model performance without PE at inference. With group size $n=8$, we vary the number of retained original prompts $m \in \{0, 1, 2, 4\}$ and evaluate GenEval performance without PE (\cref{tab:retention}). When $m=0$, the model never sees original prompts during training, resulting in degraded performance on unmodified test prompts (0.83). $m=1$ yields even worse results (0.76). We empirically find that the LM quickly discovers prompt variants more suitable for the FM, causing the single original prompt to consistently receive negative advantages and thus ineffective gradient updates. Setting $m=2$ significantly improves performance to 0.94, as multiple original prompts compete within the group, enabling meaningful advantage signals and robust learning on the original prompt distribution.

\begin{table}[t]
\centering
\caption{Ablation on prompt retention mechanism. We report GenEval scores without PE at inference with varying numbers of retained original prompts $m$ (group size $n=8$).}
\label{tab:retention}
\resizebox{0.85\columnwidth}{!}{%
\begin{tabular}{lcccccc|c}
\toprule
\textbf{Retention} &  \textbf{1 Obj.} & \textbf{2 Obj.} & \textbf{Cnt.} & \textbf{Clr.} & \textbf{Pos.} & \textbf{Attr.} & \textbf{Avg.}$\uparrow$ \\
\midrule
$m=0$ & 0.99 & 0.94 & 0.84 & 0.89 & 0.51 & 0.83 & 0.83 \\
$m=1$ & 0.98 & 0.88 & 0.83 & 0.85 & 0.37 & 0.64 & 0.77 \\
$m=2$  & \textbf{1.00} & \textbf{0.96} & \textbf{0.95} & \textbf{0.95} & \textbf{0.93} & \textbf{0.85} & \textbf{0.94} \\
$m=4$ & 1.00 & 0.96 & 0.94 & 0.88 & 0.93 & 0.81 & 0.92 \\
\bottomrule
\end{tabular}%
}
\end{table}

\section{Conclusion}
\label{sec:conclusion}

We presented PromptRL, a framework that jointly trains language models and flow-matching models within a unified reinforcement learning loop for text-to-image generation. Our approach leverages LM-generated prompt variations to expand the exploration space while simultaneously co-evolving a prompt enhancement module that improves generation quality at inference time. Extensive experiments demonstrate that PromptRL achieves state-of-the-art performance across multiple benchmarks.

\section*{Impact statement}
This paper presents work whose goal is to advance the field of joint RL training on both language models and diffusion models. There are many potential societal consequences of our work, none of which we feel must be specifically highlighted here.

\bibliography{main}
\bibliographystyle{icml2026}

\newpage
\appendix
\onecolumn
\section{Training details}

\noindent \textbf{Pseudo code.} Algorithm~\ref{alg:promptrl} presents the complete PromptRL training procedure. The algorithm alternates between generating prompt variants via the LM, producing images through the FM, and updating both models using group-normalized advantages. Key design choices include the prompt retention mechanism (lines 4--9) that maintains $m$ original prompts per group, and the selective gradient updates that train the LM only on refined prompts while the FM learns from all samples.

\begin{algorithm}[t]
\caption{PromptRL: Joint RL Training of LM and FM}
\label{alg:promptrl}
\begin{algorithmic}[1]
\REQUIRE Training prompts $\mathcal{D}$, LM $\pi_{\text{LM}}$, FM $\pi_{\text{FM}}$, reward function $R$, group size $n$, retention number $m$
\ENSURE Optimized $\pi_{\text{LM}}$ and $\pi_{\text{FM}}$

\FOR{each iteration}
    \STATE Sample batch of prompts $\{p_0^{(j)}\}_{j=1}^{B}$ from $\mathcal{D}$
    \FOR{each prompt $p_0^{(j)}$}
        \STATE \textcolor{gray}{\textit{// Generate prompt variants}}
        \STATE $\mathcal{P}^{(j)} \leftarrow \{p_0^{(j)}\}^m$ \textcolor{gray}{\quad\textit{// Retain $m$ original prompts}}
        \FOR{$i = m+1$ to $n$}
            \STATE $p_i^{(j)} \sim \pi_{\text{LM}}(\cdot \mid p_0^{(j)})$ \textcolor{gray}{\quad\textit{// LM refinement}}
            \STATE $\mathcal{P}^{(j)} \leftarrow \mathcal{P}^{(j)} \cup \{p_i^{(j)}\}$
        \ENDFOR
        \STATE \textcolor{gray}{\textit{// Generate images and compute rewards}}
        \FOR{each $p_i \in \mathcal{P}^{(j)}$}
            \STATE Sample $\boldsymbol{\epsilon}_i \sim \mathcal{N}(0, I)$
            \STATE $\mathbf{x}_i \leftarrow \pi_{\text{FM}}(\cdot \mid p_i, \boldsymbol{\epsilon}_i)$
            \STATE $r_i \leftarrow R(\mathbf{x}_i, p_i)$
        \ENDFOR
        \STATE \textcolor{gray}{\textit{// Group-wise advantage normalization}}
        \STATE $\mu^{(j)}, \sigma^{(j)} \leftarrow \text{mean}(\{r_i\}), \text{std}(\{r_i\})$
        \FOR{$i = 1$ to $n$}
            \STATE $A_i^{(j)} \leftarrow (r_i - \mu^{(j)}) / (\sigma^{(j)} + \epsilon)$
        \ENDFOR
    \ENDFOR
    \STATE \textcolor{gray}{\textit{// Update LM (only on refined prompts)}}
    \STATE $\theta_{\text{LM}} \leftarrow \theta_{\text{LM}} + \alpha_{\text{LM}} \sum_{j,i>m} A_i^{(j)} \nabla \log \pi_{\text{LM}}(p_i^{(j)} \mid p_0^{(j)})$
    \STATE \textcolor{gray}{\textit{// Update FM (on all samples)}}
    \STATE $\theta_{\text{FM}} \leftarrow \theta_{\text{FM}} + \alpha_{\text{FM}} \sum_{j,i} A_i^{(j)} \nabla \log \pi_{\text{FM}}(\mathbf{x}_i^{(j)} \mid p_i^{(j)}, \boldsymbol{\epsilon}_i)$
\ENDFOR
\end{algorithmic}
\end{algorithm}

\noindent \textbf{Training configurations.} Table~\ref{tab:training_config} summarizes the hyperparameters and computational resources used across all experiments. We adopt consistent settings where possible to ensure fair comparisons, with task-specific adjustments for learning rates and KL coefficients based on preliminary experiments. For image editing, we use higher resolution (1024$\times$1024) and fewer SDE steps to balance quality and efficiency, as discussed in \cref{sec:PromptRL}. 

\begin{table*}[t]
\centering
\caption{Training configurations and hyperparameters for PromptRL across different reward objectives. All experiments use GRPO as the base RL algorithm. ``—'' indicates not applicable or same as default.}
\label{tab:training_config}
\resizebox{\textwidth}{!}{%
\begin{tabular}{l|ccc|c|c}
\toprule
\multirow{2}{*}{\textbf{Configuration}} & \multicolumn{3}{c|}{\textbf{Text-to-Image}} & \textbf{Multi-Reward} & \textbf{Image Editing} \\
\cmidrule{2-6}
 & GenEval & OCR & PickScore &  & EditReward \\
\midrule
\multicolumn{6}{l}{\textit{Model Setup}} \\
\midrule
Base FM & FLUX.1-dev & FLUX.1-dev & FLUX.1-dev & FLUX.1-dev & FLUX.1-Kontext \\
Language Model & \multicolumn{5}{c}{Qwen2.5-VL-3B-Instruct} \\
FM Parameters & \multicolumn{4}{c|}{12B} & 12B \\
LM Parameters & \multicolumn{5}{c}{3B} \\
\midrule
\multicolumn{6}{l}{\textit{Training Details}} \\
\midrule
Training Dataset & FlowGRPO-GenEval & FlowGRPO-OCR & Pick-a-Pic & Mixed & OmniEdit (10k) \\
Training Samples & 50,000 & 19,653 & 25,432 & 95,085 & 10,000 \\
Training Resolution & $512 \times 512$ & $512 \times 512$ & $512 \times 512$ & $512 \times 512$ & $1024 \times 1024$ \\
Training Precision & bfloat16 & bfloat16 & bfloat16 & bfloat16 & bfloat16 \\
Inference Steps (Rollout) & 20 & 20 & 20 & 20 & 8 \\
SDE Steps & 20 & 20 & 20 & 10 & 4 \\
Group Size ($n$) & 8 & 8 & 8 & 8 & 8 \\
Prompt Retention ($m$) & 2 & 2 & 2 & 2 & 2 \\
Batch Size & 1 & 1 & 1 & 1 & 1 \\
K epochs & 1 & 1 & 1 & 1 & 1 \\
Learning Rate (FM) & $3{\times}10^{-7}$ & $3{\times}10^{-7}$ & $3{\times}10^{-7}$ & $10^{-7}$ & $2{\times}10^{-7}$ \\
Learning Rate (LM) & $10^{-6}$ & $10^{-6}$ & $10^{-6}$ & $10^{-6}$ & $4\times10^{-7}$ \\
Optimizer & AdamW & AdamW & AdamW & AdamW & AdamW \\
KL Coefficient~(FM)  & $4{\times}10^{-3}$ & $4{\times}10^{-3}$ & $2{\times}10^{-3}$ & $2{\times}10^{-3}$ & $10^{-2}$ \\
KL Coefficient~(LM)  & $10^{-2}$ & $10^{-2}$ & $10^{-2}$ & $10^{-2}$ & $10^{-2}$ 
\\
\midrule
\multicolumn{6}{l}{\textit{Reward Configuration}} \\
\midrule
Reward Model & GenEval & OCR Accuracy & PickScore & Tag-based & EditReward \\
Format Reward ($\lambda_{\text{Format}}$) & 1.0 & 1.0 & 1.0 & 1.0 & 1.0 \\
Generation Reward ($\lambda_{\text{Gen}}$) & 1.0 & 1.0 & 1.0 & 1.0 & 1.0 \\
Reward Normalization & Group-wise & Group-wise & Group-wise & Per-tag & Group-wise \\
\midrule
\multicolumn{6}{l}{\textit{Training Cost}} \\
\midrule
Number of GPUs & 8 & 8 & 8 & 8 & 8\\
GPU Type & H100 & H100 & H100 & H100 & H100 \\
Training Rollouts & 0.2M & 0.05M & 0.13M & 0.5M & 0.06M \\
\bottomrule
\end{tabular}%
}
\end{table*}

\section{Discussion}

\begin{table}[!b]
\centering
\caption{Generalization of PromptRL's prompt enhancer (PE) to unseen flow models on GenEval. All evaluations are conducted at 1024$\times$1024 resolution with 20 inference steps. We compare official pretrained weights without PE, with our PE, and with model-specific RL-tuned PE. Both PE modules use Qwen-2.5-VL-Instruct-3B. Our learned PE demonstrates strong generalization capability, effectively improving pretrained models that were not seen during training.}
\label{tab:generalization}
\begin{tabular}{lccc}
\toprule
\textbf{Model} & \textbf{Original} & \textbf{Ours' PE} & \textbf{Prompt-only RL} \\
\midrule
SANA & 0.62 & 0.70 & 0.76 \\
SD3 & 0.58 & 0.77 & 0.83 \\
\bottomrule
\end{tabular}
\end{table}
\noindent \textbf{Generalization of prompt enhancer to unseen FMs.}
We investigate whether the prompt enhancement module trained with PromptRL generalizes to flow models not seen during training. Specifically, we evaluate our prompt enhancer on SANA~\citep{xie2024sana} and SD3 using GenEval at 1024 resolution with 20 inference steps (\cref{tab:generalization}). Although our prompt enhancer does not match the performance of model-specific prompt enhancers trained via prompt-only RL for each FM, it significantly improves over the original model performance. For SANA, our enhancer improves GenEval from 0.62 to 0.70, and for SD3 from 0.58 to 0.77. These results demonstrate that PromptRL learns generalizable linguistic refinements that benefit diverse flow models, suggesting the LM captures semantic patterns broadly useful for T2I generation rather than overfitting to FLUX.1-dev-specific characteristics.

\noindent \textbf{Comparison to flow-only RL.} A natural question is whether the gains from PromptRL can be matched by simply scaling up flow-only RL training. To investigate this, we train a flow-only baseline with twice the number of rollouts and compare it against PromptRL in \cref{tab:flowonly}. Despite the 2$\times$ computational advantage, flow-only RL still underperforms PromptRL across all metrics (GenEval: 0.93 \vs\ 0.97, OCR: 0.93 \vs\ 0.98, PickScore: 23.85 \vs\ 24.05). This result suggests that the limitations of flow-only RL—namely exploration collapse and prompt overfitting identified in \cref{sec:und}—cannot be overcome by additional rollouts alone. Joint LM-FM optimization fundamentally reshapes the optimization landscape by continuously injecting linguistic diversity, enabling the FM to escape narrow reward modes and achieve higher performance ceilings.

\begin{table}[t]
\centering
\caption{Comparison between PromptRL and flow-only RL. Flow-only RL is trained with 2$\times$ the number of rollouts to control for computational budget.}
\label{tab:flowonly}
\begin{tabular}{lcccc}
\toprule
\textbf{Method} & \textbf{Rollouts} & \textbf{GenEval}$\uparrow$ & \textbf{OCR}$\uparrow$ & \textbf{PickScore}$\uparrow$ \\
\midrule
Flow-only RL & 2$\times$ & 0.93 & 0.93 & 23.85 \\
PromptRL     & 1$\times$ & 0.97 & 0.98 & 24.05 \\
\bottomrule
\end{tabular}%
\end{table}

\noindent \textbf{Limitations.} 
Although PromptRL achieves strong performance through joint LM-FM optimization, the FM and LM develop a degree of co-adaptation during training. Specifically, when replacing our co-trained prompt enhancer with a different LM (\eg, Qwen-3) at inference time, we observe a performance drop on GenEval from 0.97 to 0.88. This indicates that the FM becomes partially specialized to the linguistic patterns produced by its training-time LM partner. Fortunately, this co-adaptation is by design rather than a fundamental flaw—PromptRL explicitly aims to jointly optimize both components for deployment as a unified system. In practical scenarios where the co-trained LM and FM are used together, this tight coupling translates into the state-of-the-art performance we report. Future work could explore techniques such as multi-LM training or regularization strategies to further improve cross-LM generalization when broader compatibility is desired.

\noindent \textbf{Why JointRL matters?}
As T2I models become more powerful, they also become more sensitive to how prompts are phrased—small linguistic changes can lead to significant differences in generation quality. This makes the PE module just as important as the FM itself. Yet existing methods typically train these two components separately, either through independent SFT or RL. We believe this isolated approach misses a crucial point: the best way to refine prompts depends on the current FM, and vice versa. By treating PE and FM as a unified system, PromptRL allows both to improve together—the LM discovers prompt variants that yield higher rewards, while the expanded prompt diversity provides richer exploration signals that accelerate FM training efficiency.

\end{document}